\documentclass[11pt]{article} % For LaTeX2e
\usepackage{rldmsubmit,palatino}
\usepackage{amsmath, amssymb, amsthm}
\usepackage{graphicx}
\usepackage{natbib}
\theoremstyle{plain}
\newtheorem{theorem}{Theorem}[section]

\newtheorem{lemma}[theorem]{Lemma}

\theoremstyle{definition}
\newtheorem{definition}[theorem]{Definition}
\newtheorem{assumption}[theorem]{Assumption}
\theoremstyle{remark}

\title{Social Cooperation in Conversational AI Agents}

\author{
Mustafa Mert Çelikok \\
Department of Intelligent Systems\\
Delft University of Technology\\
Delft, 2600 AA, The Netherlands \\
\texttt{m.m.celikok@tudelft.nl} \\
\And
Saptarashmi Bandyopadhyay \\
Department of Computer Science\\
University of Maryland\\
College Park, MD 20742, USA \\
\texttt{saptab1@umd.edu} \\
\And
Robert Loftin \\
School of Computer Science\\
University of Sheffield \\
Sheffield, S10 2TN, UK \\
\texttt{r.loftin@sheffield.ac.uk} \\
}
\begin{document}

\maketitle

\begin{abstract}
The development of AI agents based on large, open-domain language models (LLMs) has paved the way for the development of general-purpose AI assistants that can support human in tasks such as writing, coding, graphic design, and scientific research.  A major challenge with such agents is that, by necessity, they are trained by observing relatively short-term interactions with humans.  Such models can fail to generalize to long-term interactions, for example, interactions where a user has repeatedly corrected mistakes on the part of the agent.  In this work, we argue that these challenges can be overcome by explicitly modeling humans' \emph{social intelligence}, that is, their ability to build and maintain long-term relationships with other agents whose behavior cannot always be predicted.  By mathematically modeling the strategies humans use to communicate and reason about one another over long periods of time, we may be able to derive new \emph{game theoretic} objectives against which LLMs and future AI agents may be optimized.
\end{abstract}

\keywords{
Social Intelligence, Cooperation, Conversational Agents, Imitation Learning, Reinforcement Learning, Language Models
}

\acknowledgements{Mustafa Mert Çelikok is funded by the Hybrid Intelligence Center, a 10-year programme granted by the Dutch Ministry of Education, Culture and Science through the Netherlands Organisation for Scientific Research, https://hybridintelligence-centre.nl.}

\startmain % to start the main 1-4 pages of the submission.

\section{Introduction}
\label{introduction}

%%% INTRO V3 %%%

In order to effectively collaborate with human users, conversational AI assistants~\cite{ross2023programmer,he2023recommender} must be able to sustain long-term interactions with these users, while adapting to their individual needs.  As a hypothetical example, consider a team lead, Alice, at a software firm who brings such an AI assistant into her daily workflow.  Alice invests time teaching the AI about her communication style, emphasizing empathy for colleagues and the importance of delivering constructive feedback rather than blunt directives.  Over several weeks, however, subtle yet critical issues begin to surface.  Despite repeated guidance, the AI continues to provide terse, impersonal updates in high-pressure situations.  It also allocates work items in an unpredictable and often unfair manner to colleagues who have expressed personal constraints.   Alice's team have responded by no longer expressing these constraints at all, but the AI fails to request this information when allocating critical tasks.  Alice ultimately abandons the AI altogether.

We can imagine that the AI assistant described above is based on a large language model trained or fine-tuned on a corpus of conversations between employees in similar workplaces.  This example then highlights two major technical challenges in creating such systems.  The first is the need to learn appropriate behavior conditioned on long conversation histories, that is, models that generalize well to large context windows~\cite{tworkowski2024focused}.  The failure off the AI to adhere to previous guidance, or to appropriately balance workloads, may reflect the underlying model's inability to replicate human behavior over long time horizons.  The second is that even small differences between the AI's behavior and the expected behavior of a human assistant can lead to substantial changes in the way the users interact with the system (such as intentionally withholding information).  Because this new behavior may not have be seen in data containing only human-human interactions, the AI may not have learned effective responses (like asking follow-up questions).

In this work, we examine whether such issues may be addressed by assuming that human behavior in collaborative settings is not arbitrary.  Specifically, we mathematically formalize two natural assumptions: 1) that individual humans will adapt to other agents (human or AI) over time, and 2) that human populations (e.g., technology professionals) must have preexisting "conventions" that allow individuals to reliably cooperate with new teammates.  We then develop a simplified model of the problem of learning to cooperate with such a \emph{socially intelligent} population, and provide sample complexity bounds which show that effective cooperation strategies can be learned more efficiently for populations that satisfy these assumptions.  Importantly, our theoretical results suggest that direct imitation learning (or supervised fine-tuning), may not be the most appropriate objective for training AI agents to cooperate with such human users.

\section{Preliminaries}
\label{preliminaries}

We model long-term collaborations between members of the target population as \emph{repeated two-player matrix games with private utilities.}  We denote a \emph{class} of such repeated games with a tuple $\mathcal{G} = (\mathcal{I}, \mathcal{A}, \Theta, G, T)$ where $\mathcal{I} = \{ 1,2\}$ is the set of agents, $\mathcal{A}$ is the set of $N$ pure strategies available to both agents (called \emph{actions} henceforth), $\Theta$ is a space of possible \emph{types}, $G$ is a function that maps an agent's type $\theta \in \Theta$ to a payoff matrix $G(\theta) \in \mathbb{R}^{N \times N}$, and $0< T < \infty$ is a fixed number of stages. Let $\boldsymbol{\theta} = (\theta_1, \theta_2)$ denote a joint type for both agents. Then, a specific instance of a game from this class is given by $\mathcal{G}(\boldsymbol{\theta}) = (\mathcal{I}, \mathcal{A}, G(\boldsymbol{\theta}), T)$ such that $G(\boldsymbol{\theta}) = [G(\theta_1), G(\theta_2)^{\top}]$ is its payoff matrix.  Intuitively, we can think of an agent's type as capturing their preferences for how collaborative tasks should be accomplished.

In a single \textit{episode}, the agents play $\mathcal{G}(\boldsymbol{\theta})$ for $T$ stages. We let $a^{1}_t$ and $a^{2}_t$ denote the actions chosen by agents 1 and 2 in stage $0 < t \leq T$.  For mixed strategies $\sigma, \sigma' \in \Delta(\mathcal{A})$, we let $G(\sigma, \sigma' ; \theta_i) = \sigma^{\top} G(\theta_i) \sigma'$.  We overload $a^{1}_t$ and $a^{2}_t$ to also denote the mixed strategies that assign all probability mass to actions $a^{1}_t$ and $a^{2}_t$, such that $G(a^{1}_t, a^{2}_t; \theta_1)$ and $G(a^{1}_t, a^{2}_t; \theta_2)$ are agent $1$ and $2$'s realized payoffs at stage $t$.  We also assume, without the loss of generality, that for all $\theta \in \Theta$, $G(a^1_t = a, a^2_t = a', \theta) \in [0, 1], \forall a,a' \in \mathcal{A}$. In other words, payoffs are always bounded in $[0,1].$   

Let $\mathcal{H}_t = (\mathcal{A} \times \mathcal{A})^t$ be the set of histories of length $t$ (with $\mathcal{H}_0 = \{ \emptyset \}$), and let $\mathcal{H}_{\leq t} = \bigcup^{t}_{s=0}\mathcal{H}_s$ be the set of all histories of length at most $t$.  The \emph{meta-strategy} space $\Pi$ for an agent is then the space of mappings $\pi : \Theta \times \mathcal{H}_{\leq T-1} \mapsto \Delta(\mathcal{A})$, where $\Delta(\mathcal{A})$ is the set of probability distributions over the action set.  As a functional of types, a meta-strategy $\pi(\theta, \cdot)$ maps a type $\theta$ to a \textit{behavioral strategy}~\citep[Chapter~5.2.2]{shoham2008multiagent} that maps histories of play to action distributions, such that $a^{i}_t \sim \pi_i (\theta_i, h_{t-1})$.  We denote agent $i$'s expected total payoff for following meta-strategy $\pi$ against $\pi'$ as $M_i (\pi, \pi' ; \theta, \theta') = \text{E}[ \sum^{T}_{t=1} G(a^{i}_t, a^{-i}_t ; \theta_i) \ | \ 
     \pi_i \!=\! \pi, \pi_{-i} \!=\! \pi', \nonumber 
     \theta_i \!=\! \theta, \theta_{-i} \!=\! \theta']$ where the expectation is over the strategies.

\section{Social Intelligence}
\label{social_intelligence}

As described in section 1, we assume that the members of the population are individually rational and can cooperate effectively with each other. We formalize these two assumptions via \emph{consistency} and \emph{compatibility}. In essence, consistency requires an agent $i$ to achieve bounded external regret regardless of its type or partner. Also called the no-regret property, this requirement is often seen as necessary for individual rationality. Compatibility requires that when paired with another member of the population, with high-probability the agents will achieve the same expected utility as they would under some Pareto-efficient equilibrium strategy of the underlying stage game. Similar requirements have been used previously to define successful cooperation~\citep{powers2004targeted}.

\begin{definition}[Consistency]
\label{def:consistency}
Let $R^{\text{ext}}_i (h ; \theta_i) = \max_{a^{i} \in [N]} \sum^{\vert h \vert}_{t=1} \left\{ G(a^{i}, a^{-i}_t(h) ; \theta_i) - G( a^{i}_t(h), a^{-i}_t (h) ; \theta_i) \right\}$ denote the external regret for agent $i$. For $\delta, \epsilon, T > 0$, an agent $i \in \{1,2\}$ is $(\delta, \epsilon, T)$-\textit{consistent} if, for all types $\theta \in \Theta$, and \textit{any} partner strategy, we have that $\frac{1}{T} R^{\text{ext}}_i (h_T ; \theta) \leq \epsilon$ with probability at least $1-\delta$.
\end{definition}

Let $\mathcal{N}(\boldsymbol{\theta}) \subseteq \Delta(\mathcal{A}) \times \Delta(\mathcal{A})$ be the set of Nash equilibria (NE) of $\mathcal{G}(\boldsymbol{\theta})$, and $\mathcal{P}(\boldsymbol{\theta}) \subseteq \mathcal{N}(\boldsymbol{\theta})$ denote the set of Pareto optimal Nash equilibria. In this work, we say that a strategy profile $\langle \sigma_1, \sigma_2 \rangle \in \mathcal{P}(\boldsymbol{\theta})$ if and only if $\langle \sigma_1, \sigma_2 \rangle  \in \mathcal{N}(\boldsymbol{\theta})$, and there does not exist $\langle \sigma'_1, \sigma'_2 \rangle  \in \mathcal{N}(\boldsymbol{\theta})$ such that $G(\sigma'_1, \sigma'_2; \theta_1) > G(\sigma_1, \sigma_2; \theta_1)$ \textit{and} $G(\sigma'_2, \sigma'_1; \theta_2) > G(\sigma_2, \sigma_1; \theta_2)$.  This means that $\langle \sigma_1, \sigma_2 \rangle$ is a PONE if it is a Nash equilibrium of $\mathcal{G}(\boldsymbol{\theta})$, and it is not \textit{strongly} Pareto-dominated by any other Nash equilibrium of $\mathcal{G}(\boldsymbol{\theta})$. Intuitively, if two agents are individually consistent, and willing to cooperate with each other, their joint payoff profile should come close to a PONE. We formalize this intuition as follows, which is the approximate and finite-horizon version of the one provided in~\citet{powers2004targeted}.

\begin{definition}[Compatibility]
\label{def:compatibility}
    For $\delta, \epsilon, T > 0$, two agents $\pi^1$ and $\pi^2$ are $(\delta, \epsilon, T)$-\textit{compatible} if, when played together, for any joint type $\boldsymbol{\theta}$, w.p. at least $1-\delta$, $\exists \langle \sigma^{*}_1, \sigma^{*}_2 \rangle \in \mathcal{P}(\boldsymbol{\theta})$ s.t. $\frac{1}{T}\sum^{T}_{t=1}  G(\sigma^{*}_{i}, \sigma^{*}_{-i} ; \theta_i) - G(a^{i}_t, a^{-i}_t ; \theta_i) \leq \epsilon,$ for both $i=1$ and $i=2$.
\end{definition}

We argue that it is natural to model an existing population of cooperating agents as a set of approximately compatible, but otherwise heterogeneous agents. We therefore introduce the more general idea of a socially intelligent \textit{class} of agents that are compatible with any other member of their class:

\begin{definition}[Social Intelligence]
\label{def:social_intelligence}
A set $\mathcal{C}$ of agents forms a \textit{socially intelligent class} w.r.t. $\Theta$ if, for some $\delta, \epsilon, T > 0$, each agent $\pi \in \mathcal{C}$ is $(\delta, \epsilon, T)$-consistent for all $\theta \in \Theta$, and any two agents $\pi, \pi' \in \mathcal{C}$ are $(\delta, \epsilon, T)$-compatible over all joint types $\Theta$. An individual agent $\pi$ is called \textit{socially intelligent} if it forms a socially intelligent class $\{ \pi \}$ with itself.
\end{definition}

To see the implications of SI for communication, consider the case when a cautious employee is paired with an employee who needs and values directness. If they belong to $\mathcal{C},$ they should be able to identify each other's types and adapt their communication strategies to avoid individual regret and achieve approximate compatibility. Our goal is to choose a meta-strategy for the AI agent that can communicate and cooperate with a partner drawn from some \textit{target population} nearly as effectively as agents from this population do so with one another. For the class of games $\mathcal{G} = (\mathcal{I}, \mathcal{A}, \Theta, G, T)$, we will let the target population be a set $\mathcal{C}$ of agents forming a $(\delta,\epsilon,T)$-SI class with respect to $\Theta$.  

Ideally, we would hope to choose an AI meta-strategy $\pi$ that can cooperate with $\mathcal{C}$ \textit{without} any additional information about the strategies in $\mathcal{C}$. We therefore consider the problem of learning a cooperative meta-strategy using prior observations of members of the target population interacting with one another.  We define a \textit{social learning problem} by a tuple $\{ \mathcal{G}, \mathcal{C}, \rho, \mu \}$, where $\mathcal{C}$ is the target population (SI w.r.t. $\Theta$), $\rho$ is a distribution over $\mathcal{C}$, while $\mu$ is a distribution over the joint type space $\Theta \times \Theta$.  We can think of $\mathcal{C}$ as the set of possible strategies that any member of the target population might follow, while $\rho$ is the frequency of those strategies within the population.  To choose an AI strategy, we leverage a dataset $\mathcal{D} = \{(\theta^j_1, \theta^j_2, h^j_T) | j \in [n] \}$ covering $n$ \textit{episodes} of length $T$.  In each episode $j$, two agents $\pi^1_j$ and $\pi^2_j$ are sampled independently from $\rho$, and played together under the joint type $\boldsymbol{\theta}^j \sim \mu$.  The AI agent observes the full history $h^j_T$, along with the agents' types $\theta^j_1$ and $\theta^j_2$.  We denote a specific learning algorithm as a data conditioned strategy $\pi(\mathcal{D})$.

We seek an AI strategy that minimizes the regret relative to some Pareto optimal solution to $G(\boldsymbol{\theta})$.  Rather than minimizing regret in terms of the AI's own payoffs, however, we seek to minimize \textit{partner's} relative to their (worst case) PONE in $G(\boldsymbol{\theta})$.  We formalize this regret with the following definition:

\begin{definition}[Altruistic Regret]
\label{def:altruistic_regret}
Let the $(\sigma^*_i, \sigma^*_{-i})$  denote the PONE with the \emph{lowest payoff} for the agent $-i$ where $i \in \{1,2\}$. The altruistic regret of agent $i$ is defined as: $R^{\text{alt}}_i(h_T; \theta_{-i})=\sum^{T}_{t=1}  G(\sigma^*_i, \sigma^*_{-i}; \theta_{-i}) - G(a^{i}(h_t), a^{-i}(h_t) ; \theta_{-i}).$

\end{definition}

In practical cooperation tasks, we would expect outcomes that have low regret for the partner will have low regret for the AI agent as well. The cooperation objective for the AI agent can then be formalized as minimizing its \textit{expected} altruistic regret over partners sampled from $\rho$ and types sampled from $\mu$. Unlike the definition suggests, the AI agent must know its own type as well. This is due to the fact that, if the AI agent fails to imitate a human of its type or fail to communicate its type correctly, the partner might switch to a 
safe strategy. This safe strategy can involve permanently disengaging the AI agent, as described in our introductory example.

\subsection{Theoretical Ablations}

Even though it is natural to model populations through our definition of social intelligence, it is not clear whether we need both consistency and compatibility. Let us assume that $\mathcal{C}$ consists of agents that are consistent but not necessarily compatible. The most general class in this case is the class of all no-external-regret learners (no-regret henceforth). 
The long-run average of no-regret learning converges to the set of coarse correlated equilibria. The question is whether the AI agent can learn to do better than a coarse correlated equilibrium (CCE) when paired with a member of $\mathcal{C}$, using only a dataset $\mathcal{D}$ that consists of histories of play for different CCE.

Unfortunately, we cannot guarantee learning in the case of consistency alone. The theorem 3 of \citet{monnot2017limits} shows that given any coarse correlated equilibrium of a two-player normal-form game, there exists a pair of no-regret learners that would converge to it. Since $\mathcal{C}$ can be any subset of no-regret learners, we cannot exclude those who converge to inefficient CCE. If the class $\mathcal{C}$ contains only the agents that converge to Pareto-inefficient CCE, we cannot hope to learn optimal strategies from any dataset with respect to the altruistic regret. Additionally, regardless of the dataset, when the AI's strategy is deployed, it will face a partner drawn from $\mathcal{C}$ whose type is unknown. We may hope to learn a classifier to quickly infer our partner's type online from their behavior, assuming there exists a mapping from initial behavior to types. However, since $\mathcal{C}$ consists only of no-regret learners guaranteed to converge to a CCE in self-play, they have no reason to communicate their types to each other.

Now let us assume that the members of $\mathcal{C}$ are compatible but not consistent. Imagine that the members of $\mathcal{C}$ have an established handshake protocol, which allows them to communicate and identify each other's types at the beginning of their interaction. When they identify each other after the handshake phase, they proceed with playing the agreed-upon PONE. However, if at any moment one plays the wrong action, there is no constraint on what strategy the other will switch to. 
The members of $\mathcal{C}$ can employ grim-trigger strategies that punish the other agent for the rest of the horizon, triggered by a mistake at any point. The finite-horizon setting means even with a forgiving strategy, mistakes can in effect lead to grim-trigger outcomes. Then the outsider must learn to imitate at least one member of $\mathcal{C}$ perfectly from the dataset. Therefore the offline problem in this setting reduces to imitation learning, in particular the no-interaction case from \citet{rajaraman2020toward}. Unfortunately, the sample complexity of the no-interaction imitation learning scale exponentially with horizon in our case, since the strategies are history-dependent. 

\subsection{Upper Bound for Social Cooperation}
A key idea behind this work is that against a socially intelligent target population, rather than trying to imitate a member of the population perfectly throughout the entire episode, the AI agent only needs to imitate them long enough to learn about its partner's private type.  Once it has this information, the AI agent can leverage the fact that the partner's strategy is consistent against \textit{any} strategy, and try to ``coerce'' the human partner into playing a strategy that minimizes the altruistic regret.  We will refer to such strategies as \textit{imitate-then-commit} (IC) strategies, which use the previous observations $\mathcal{D}$ to learn an imitation strategy to follow over the first $\tilde{T} < T$ steps of the interaction.  In this section we provide an upper bound on the altruistic regret of a specific (IC) strategy, as a function of the number of episodes in $\mathcal{D}$, subject to the following assumptions:

\begin{assumption}
    \label{asm:social_intelligence}
    For $\delta, \epsilon > 0$, and some $\tilde{T} < T$, we have that (1) $\rho$ is $(\delta, \epsilon, T)$-consistent, and (2) $\rho$ is $(\delta, \epsilon, \tilde{T})$-compatible.
\end{assumption}

\paragraph{Imitation learning.}
Under an imitate-then-commit strategy, the sample complexity is defined entirely by the number of episodes the AI agent needs to observe to learn a good $\tilde{T}$-step imitation policy.  Fortunately, imitation learning is a well-studied problem, and we can largely leverage existing complexity bounds.  The one caveat is that in this setting we need bounds on the total variation distance between the distribution over the partial history $h_{\tilde{T}}$ under the population strategy $\rho$, and that under the learned strategy.  Given the dataset $\mathcal{D}$, we define the imitation strategy $\hat{\pi}^{1}_{\tilde{T}}(\mathcal{D})$ such that $\hat{\pi}^{1}_{\tilde{T}}(h; \theta, \mathcal{D})$ is the empirical distribution over agent $1$'s actions for each history-type pair $(h, \theta)$ occurring in $\mathcal{D}$, while $\hat{\pi}^{1}_{\tilde{T}}(h; \theta, \mathcal{D})$ is the uniform distribution over $N$ for $(h, \theta) \notin \mathcal{D}$.  We then define the \textit{marginal} strategy $\hat{\pi}^{1}_{\tilde{T}}$, which can be implemented by sampling a dataset $\mathcal{D}$, and then following the imitation strategy defined by $\mathcal{D}$ for the next $\tilde{T}$ steps.  We then have the following bound on the distribution of $h_{\tilde{T}}$ under the imitation strategy:

\begin{lemma}
    \label{lem:imitation_learning}
    Let $p_{\tilde{T}}$ be the distribution over partial histories $h_{\tilde{T}}$ under the population strategy $\rho$, and let $\hat{p}_{\tilde{T}}$ be their distribution under $\hat{\pi}^{1}_{\tilde{T}}$.  We have that
    \begin{equation}
        \Vert p_{\tilde{T}} - \hat{p}_{\tilde{T}} \Vert_{\text{TV}} \leq \min\left\{ \tilde{T}, \frac{N^{2(\tilde{T}+1)} \vert\Theta\vert \tilde{T}^2\log(K)}{K} \right\}, \quad \text{where } K = \vert \mathcal{D} \vert.
    \end{equation}
\end{lemma}

This bound follows directly from that of~\citet{rajaraman2020toward} via Lemma 1 of~\citet{ciosek2022imitation} (see supplementary material section 2.1 for full proof).

%%%%%%%%%%%%%%%%%%%%%%%%%%%%%%%%%%%%%%%%%%%%%%%%%%%%%%%%

\paragraph{Imitate-then-commit strategy.}
For history $h_{\tilde{T}} \in \mathcal{H}_{\tilde{T}}$, we let $\hat{z}(h_{\tilde{T}}) \in \Delta(N \times N)$ denote the empirical \textit{joint} strategy played up to and including step $\tilde{T}$.  We show that, using $\hat{z}(h_{\tilde{T}})$, it is possible to construct a \textit{mixture} $\nu$ over mixed strategies $x \in \Delta(N)$ that, in expectation over $\nu$, the partner's payoff under their best response to $x \sim \nu$ will be at least as large as their payoff under $\hat{z}(h_{\tilde{T}})$.  The corresponding IC strategy will operate as follows:
\begin{enumerate}
    \item Sample $\mathcal{D}$ and compute the imitation strategy $\hat{\pi}^{1}_{\tilde{T}}(\mathcal{D})$.
    \item Play $\hat{\pi}^{1}_{\tilde{T}}(\mathcal{D})$ for the first $\tilde{T}$ steps, and observe $h_{\tilde{T}}$.
    \item Compute a suitable mixture $\nu$ from $\hat{z}(h_{\tilde{T}})$, and sample $x \sim \nu$
    \item Sample actions from $x$ for the remaining $T - \tilde{T}$ steps.
\end{enumerate}
We then have the following upper bound on the altruistic regret achievable with an imitate-then-commit strategy:

\begin{theorem}
    \label{thm:upper_bound}
    Given that Assumption~\ref{asm:social_intelligence} holds for $\rho$, there exists a data-dependent strategy $\pi^{\text{IC}}(\mathcal{D})$ such that when played by the AI as agent $2$, the altruistic regret satisfies
    \begin{equation}
        \label{eqn:upper_bound}
        \text{E}\left[ R^{\text{alt}}_{1}(h_T, \theta_{2}) \right] \leq 2\delta + \delta(K) + \left( 2\frac{T - \tilde{T}}{T} + 1\right) \epsilon, \quad         \delta(K) = \min\left\{ \tilde{T}, \frac{N^{2(\tilde{T}+1)} \vert\Theta\vert \tilde{T}^2\log(K)}{K} \right\}       
    \end{equation}
    where $K = \vert \mathcal{D} \vert$ and the expectation is taken over $h_T$, $\boldsymbol{\theta}$, and $\mathcal{D}$.
\end{theorem}

\paragraph{Proof sketch:} By Lemma~\ref{lem:imitation_learning}, we can learn an imitation strategy such that the corresponding distribution over $h_{\tilde{T}}$ and $\hat{z}(h_{\tilde{T}})$ is close to that under $\rho$ in self-play.  As $\rho$ is compatible, both agents' payoffs under $\hat{z}(h_{\tilde{T}})$ must be close to those under \textit{some} PONE.  Finally, we can construct a mixture $\nu$ for agent $1$ such that agent $2$'s payoffs under its (approximate) best-response are almost as large as those under $\hat{z}(h_{\tilde{T}})$ (see supplementary material section~2.2).

\section{Conclusion}
\label{conclushion}

Training conversational AI agents to maintain long-term interactions and cooperation with human partners is a challenging problem. Naively reducing this problem to imitation learning leads to infeasibility, due to the statistical limitations when state spaces are too large and horizons are long. Instead, we provide formal guarantees for successful and reliable cooperation of AI agents with populations of socially intelligent agents. We present a novel definition of social intelligent populations based on the assumptions that 1) members of the population are individually rational, and 2) pairs of members can achieve performance comparable to a Pareto-optimal Nash equilibrium. We formalize the notion of consistency and compatibility of agents in repeated, two-player, general-sum matrix games with private types. 

Our theoretical guarantees are in the offline cooperation setting where the agent has to cooperate with unseen partners in the population to strategize against and manipulate no-regret policies for which we formalize the idea of altruistic regret. We provide upper bounds on the sample complexity needed to learn a successful cooperation strategy. These complexity analysis and formally proven bounds can be helpful to sustainably model the alignment problem of AI agents.

\bibliography{rldm_references}

\begin{thebibliography}{8}
\providecommand{\natexlab}[1]{#1}
\providecommand{\url}[1]{\texttt{#1}}
\expandafter\ifx\csname urlstyle\endcsname\relax
  \providecommand{\doi}[1]{doi: #1}\else
  \providecommand{\doi}{doi: \begingroup \urlstyle{rm}\Url}\fi

\bibitem[Ciosek(2022)]{ciosek2022imitation}
K.~Ciosek.
\newblock Imitation learning by reinforcement learning.
\newblock In \emph{International Conference on Learning Representations}, 2022.

\bibitem[He et~al.(2023)He, Xie, Jha, Steck, Liang, Feng, Majumder, Kallus, and McAuley]{he2023recommender}
Z.~He, Z.~Xie, R.~Jha, H.~Steck, D.~Liang, Y.~Feng, B.~P. Majumder, N.~Kallus, and J.~McAuley.
\newblock Large language models as zero-shot conversational recommenders.
\newblock In \emph{Proceedings of the 32nd ACM international conference on information and knowledge management}, pages 720--730, 2023.

\bibitem[Monnot and Piliouras(2017)]{monnot2017limits}
B.~Monnot and G.~Piliouras.
\newblock Limits and limitations of no-regret learning in games.
\newblock \emph{The Knowledge Engineering Review}, 32:\penalty0 e21, 2017.

\bibitem[Powers and Shoham(2004)]{powers2004targeted}
R.~Powers and Y.~Shoham.
\newblock New criteria and a new algorithm for learning in multi-agent systems.
\newblock \emph{Advances in Neural Information Processing Systems}, 17, 2004.

\bibitem[Rajaraman et~al.(2020)Rajaraman, Yang, Jiao, and Ramchandran]{rajaraman2020toward}
N.~Rajaraman, L.~Yang, J.~Jiao, and K.~Ramchandran.
\newblock Toward the fundamental limits of imitation learning.
\newblock \emph{Advances in Neural Information Processing Systems}, 33:\penalty0 2914--2924, 2020.

\bibitem[Ross et~al.(2023)Ross, Martinez, Houde, Muller, and Weisz]{ross2023programmer}
S.~I. Ross, F.~Martinez, S.~Houde, M.~Muller, and J.~D. Weisz.
\newblock The programmer’s assistant: Conversational interaction with a large language model for software development.
\newblock In \emph{Proceedings of the 28th International Conference on Intelligent User Interfaces}, pages 491--514, 2023.

\bibitem[Shoham and Leyton-Brown(2008)]{shoham2008multiagent}
Y.~Shoham and K.~Leyton-Brown.
\newblock \emph{Multiagent systems: Algorithmic, game-theoretic, and logical foundations}.
\newblock Cambridge University Press, 2008.

\bibitem[Tworkowski et~al.(2024)Tworkowski, Staniszewski, Pacek, Wu, Michalewski, and Mi{\l}o{\'s}]{tworkowski2024focused}
S.~Tworkowski, K.~Staniszewski, M.~Pacek, Y.~Wu, H.~Michalewski, and P.~Mi{\l}o{\'s}.
\newblock Focused transformer: Contrastive training for context scaling.
\newblock \emph{Advances in Neural Information Processing Systems}, 36, 2024.

\end{thebibliography}
\end{document}